\documentclass[letterpaper, 10 pt, conference]{ieeeconf}  

\IEEEoverridecommandlockouts                              

\usepackage{booktabs} 
\usepackage{makecell} 
\usepackage{mathtools}
\DeclareMathOperator*{\argmin}{arg\,min}
\usepackage{amsmath}   
\usepackage{amssymb}   
\DeclareMathOperator*{\argmax}{arg\,max} 
\usepackage{graphicx}
\usepackage{url}
\usepackage{eso-pic}
\urldef{\CMARPublicationDOI}\url{https://doi.org/10.1109/ICRA57385.2026.11697035}

\title{\LARGE \bf
CMAR-search: Commonsense and Memory Augmented Reasoning for Object Search in Dynamic Interactive Environments
}

\author{Kaiyao Liao$^{1\dagger}$, Qingfeng Li$^{1\dagger}$, Xinlei Zhang$^{1}$, Chen Chen$^{2}$, Qing Sun$^{2}$, Jianwei Niu$^{1*}$%
\thanks{$^{1}$School of Computer Science and Engineering, Beihang University}%
\thanks{$^{2}$Zhejiang Key Laboratory of Industrial Big Data and Robot Intelligent Systems, Hangzhou Innovation Institute of Beihang University}%
\thanks{$^{\dagger}$ These authors contributed equally to this work}%
\thanks{$^{*}$ Corresponding author}%
}

\begin{document}

\maketitle
\thispagestyle{empty}
\pagestyle{empty}

\AddToShipoutPictureFG*{%
  \AtPageLowerLeft{%
    \put(\LenToUnit{0.75in},\LenToUnit{0.12in}){%
      \parbox[b]{7in}{\fontsize{7}{8}\selectfont\raggedright
        Author's accepted manuscript. Published in: \emph{2026 IEEE International Conference on Robotics and Automation (ICRA)},
        2026, pp.~18586--18593. DOI: \CMARPublicationDOI.\par\vspace{2pt}
        \copyright~2026 IEEE. Personal use of this material is permitted. Permission from IEEE must be obtained for all other uses,
        in any current or future media, including reprinting/republishing this material for advertising or promotional purposes,
        creating new collective works, for resale or redistribution to servers or lists, or reuse of any copyrighted component
        of this work in other works.
      }%
    }%
  }%
}

\begin{abstract}
Dynamic interactive object search in large-scale human environments presents substantial challenges for existing methods. Current scene representations like 3D Scene Graphs (3DSG) only provide coarse-grained spatial segmentation and cannot identify functional areas such as storage or leisure areas. Without functional area understanding, existing methods are constrained to exhaustive sequential exploration at large scales, resulting in inefficient search behaviors—particularly in open-layout environments with numerous interactive objects such as drawers and cabinets. Moreover, these methods lack adaptability to environmental dynamics such as object relocations. To address these limitations, this paper proposes CMAR-search, a novel framework built upon Commonsense and Memory Augmented Reasoning (CMAR). Our approach leverages commonsense about area functionalities and aggregates environmental memory to construct a Functional 3D Scene Graph (F3DSG), which organizes the environment into functional areas with their associated containers. Through this structured representation, CMAR enables hierarchical action planning at both macro-area and micro-container levels, empowering the system to efficiently identify and inspect semantically relevant areas for effective object search. Notably, CMAR continuously integrates real-time perception, accumulated memory, and commonsense to dynamically relocalize objects in changing environments. Extensive experiments in simulation and real-world settings demonstrate that CMAR-search significantly surpasses state-of-the-art baselines in both success rate and search efficiency for object search in dynamic interactive environments.

\end{abstract}

\section{INTRODUCTION}

In human-centered environments, interactive object search is a fundamental embodied AI task where a robot must locate target objects through navigation and physical interaction such as opening cabinets and drawers. This problem becomes particularly challenging in large-scale, dynamic settings that demand long-horizon reasoning and semantic understanding of spatial layouts. Current approaches typically extract environmental information into 3D Scene Graphs (3DSG) that structurally represent rooms and objects \cite{moma,3DSG1,3DSG2,3DSG3,3DSG4} and utilize Large Language Models (LLMs) for high-level reasoning of robotic actions \cite{LLM1,LLM2,LLM3,ZL}. However, as environmental complexity increases, these methods encounter several critical issues: (1) In open-layout home environments, existing scene representation methods struggle to effectively segment functional areas, making it difficult to narrow down possible object search ranges; (2) their reliance on atomic actions—such as opening only one container at a time—considerably slows down search processes when handling multiple containers; and (3) they lack robust adaptation to environmental dynamics, showing limited integration of historical memory and object commonsense for efficient object relocation, which significantly restricts their real-world applicability.

   \begin{figure}[t]
      \centering
      \includegraphics[width=\linewidth, keepaspectratio]{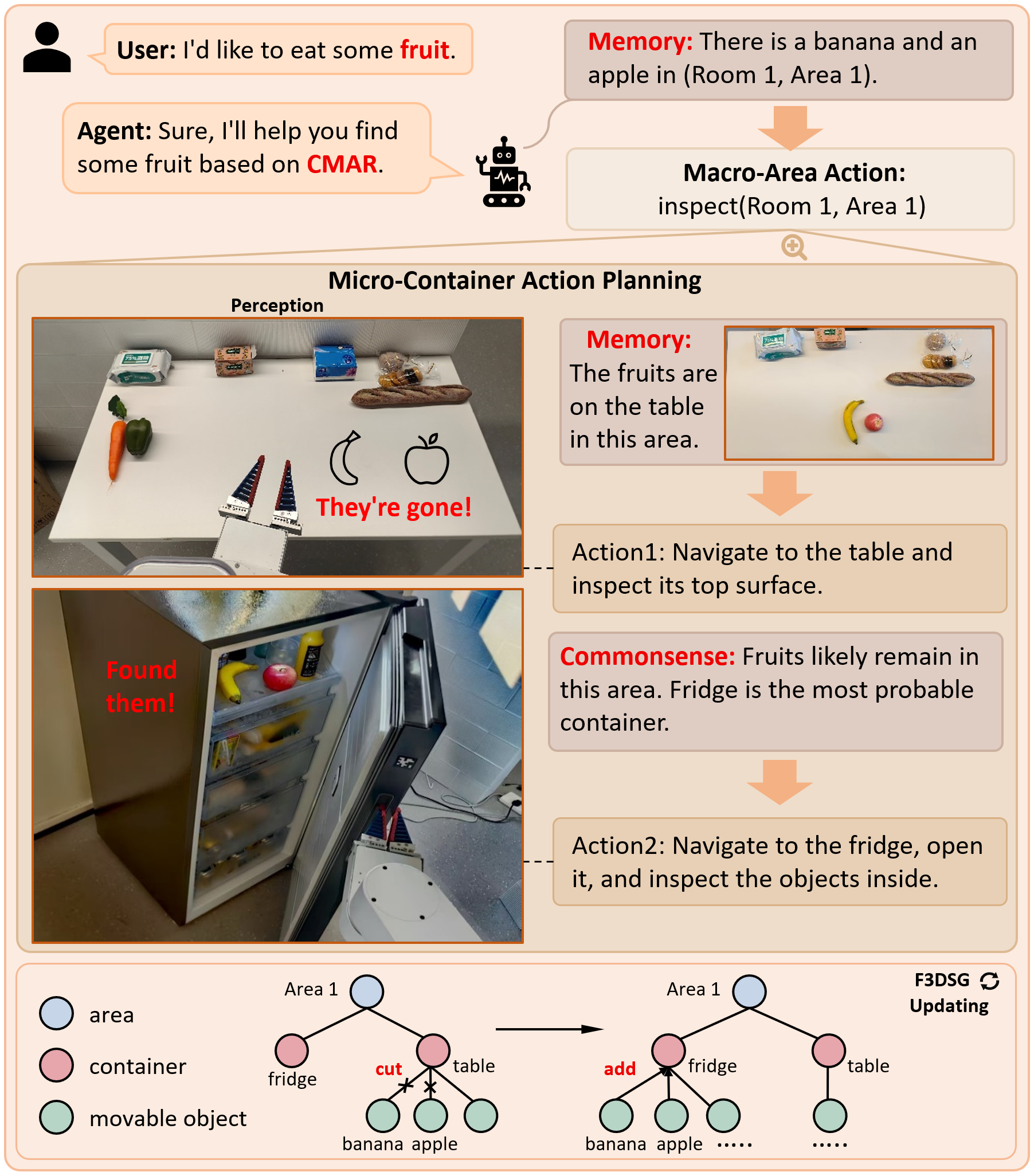}
      \caption{Our CMAR-search framework empowers the agent to perform hierarchical action planning using the F3DSG, enabling rapid object relocation in dynamic environments.}
      \label{fig:shouye}
   \end{figure}

Inspired by Kahneman’s dual-system theory of cognition (System 1 for intuitive processing and System 2 for deliberate reasoning) \cite{dual-system}, this paper proposes a novel framework centered on Commonsense and Memory Augmented Reasoning (CMAR). Our approach addresses the above limitations through two core innovations. First, in decision granularity, our Functional 3D Scene Graph (F3DSG), constructed using object-functional commonsense and spatial relationships, enables LLMs to perform macro-level planning using functional areas as units. The system directs LLMs to generate comprehensive search commands for entire areas at once, significantly reducing the number of LLMs calls and mitigating error accumulation. At the micro-level, these commands are decomposed into optimized sub-steps for inspecting all containers. Our CMAR-based sequence—considering distance, semantic similarity, and risk factors—determines the optimal execution order, further enhancing search efficiency. Second, in dynamic response, the CMAR mechanism synergistically retrieves historical memory and object-functional commonsense to infer optimal search strategies. This framework allows the robot to systematically explore new environments and adapt to changes in a human-like manner, leading to significant improvements in success rate and efficiency in dynamic interactive object search tasks.

To summarize, our main contributions are as follows:
\begin{itemize}


\item This paper proposes CMAR-search, a novel framework for dynamic interactive object search, which introduces CMAR to enable adaptation to environmental dynamics (e.g., object relocations) by synergistically leveraging real-time perception, historical memory, and commonsense, significantly improving search robustness in changing environments.

\item The framework constructs an F3DSG that organizes environments into functional areas with their associated containers, enabling hierarchical action planning at both macro-area and micro-container levels to efficiently identify and inspect semantically relevant areas, reducing unnecessary exploration in large-scale environments.

\item Extensive experiments in both simulated and real-world environments demonstrate that the proposed method significantly outperforms state-of-the-art baselines in success rate and efficiency for dynamic interactive object search tasks.

\end{itemize}

\section{RELATED WORKS}

\subsection{3DSG-based Environmental Representation}
3DSG as a sparse environmental representation can abstract hierarchical, object-centric models of environments from dense perceptual data, thereby providing a powerful interface for high-level reasoning and task planning in robotics \cite{3DSG3,3DSG4,3DSG6}. Existing research has constructed hierarchical graphs such as ``building–floor–room–object'' and enhanced their representational capacity by incorporating object-object relationship edges \cite{3DSG5,3DSG9,3DSG10}, integrating Voronoi graphs for navigation support \cite{3DSG8}, or distinguishing between static and dynamic objects \cite{3DSG4}. There are also some methods that try to implement dynamic updates for 3DSG \cite{3DSG1,zxl}. However, most of these methods rely on a pre-built, static scene graph representation, which inherently lacks the ability to dynamically evolve based on interactive experiences (e.g., inspecting container interiors) and fails to excavate the semantically critical layer of ``functional areas'' essential for task planning. In contrast, our work aims to construct a functional and dynamically updatable 3D scene graph, whose core innovation lies in leveraging LLMs to infer functional area segmentation beyond purely geometric space, thereby laying the foundation for efficient interactive object search.

\subsection{LLM-driven Robotic Planning}
LLMs have been widely used to generate behavioral sequences for robots, achieving remarkable progress in tabletop manipulation and vision-language navigation (VLN) tasks \cite{18,19}. Some studies have attempted to apply LLMs to more challenging mobile manipulation tasks in home environments, such as generating plans through information retrieval for task matching \cite{19}, fine-tuning LLMs to encode object relationships \cite{17}, or iteratively pruning scene graphs \cite{3DSG4}. MoMa-LLM \cite{moma} encodes scene graphs and instructions into natural language inputs for LLMs to generate next-step actions, enabling a certain level of interactive environmental information utilization. However, existing methods generally suffer from limited abstraction levels and single-grained instruction generation, making it difficult to adapt to the complex and diverse interactive reasoning demands in large-scale environments. This study significantly enhances the reasoning efficiency and generalization capability of LLMs in interactive tasks by introducing a hierarchically abstracted action space and structured task representation.

\subsection{Object Search}
Traditional object search methods include heuristic approaches based on frontier exploration \cite{21} and reinforcement learning models \cite{22}. With the increasing application of scene graphs, graph neural networks (GNNs) have been used to search for objects with relational constraints within graphs \cite{24} or to cope with dynamic changes \cite{25,26}. Recent studies have employed LLMs to assess semantic similarity for guiding search \cite{27,28}, but these are still limited to non-interactive navigation tasks. Schmalstieg et al. \cite{7} explicitly proposed the interactive object search task, emphasizing operations on containers to find interior objects. MoMa-LLM \cite{moma} further supports interactive search using scene graphs but lacks adaptability to environmental changes. We propose the CMAR mechanism, integrating real-time perception, historical memory, and functional commonsense, to enable efficient and robust interactive object search in dynamic environments.

\section{METHOD}
   \begin{figure*}[t]
      \centering
      \includegraphics[width=\linewidth, keepaspectratio]{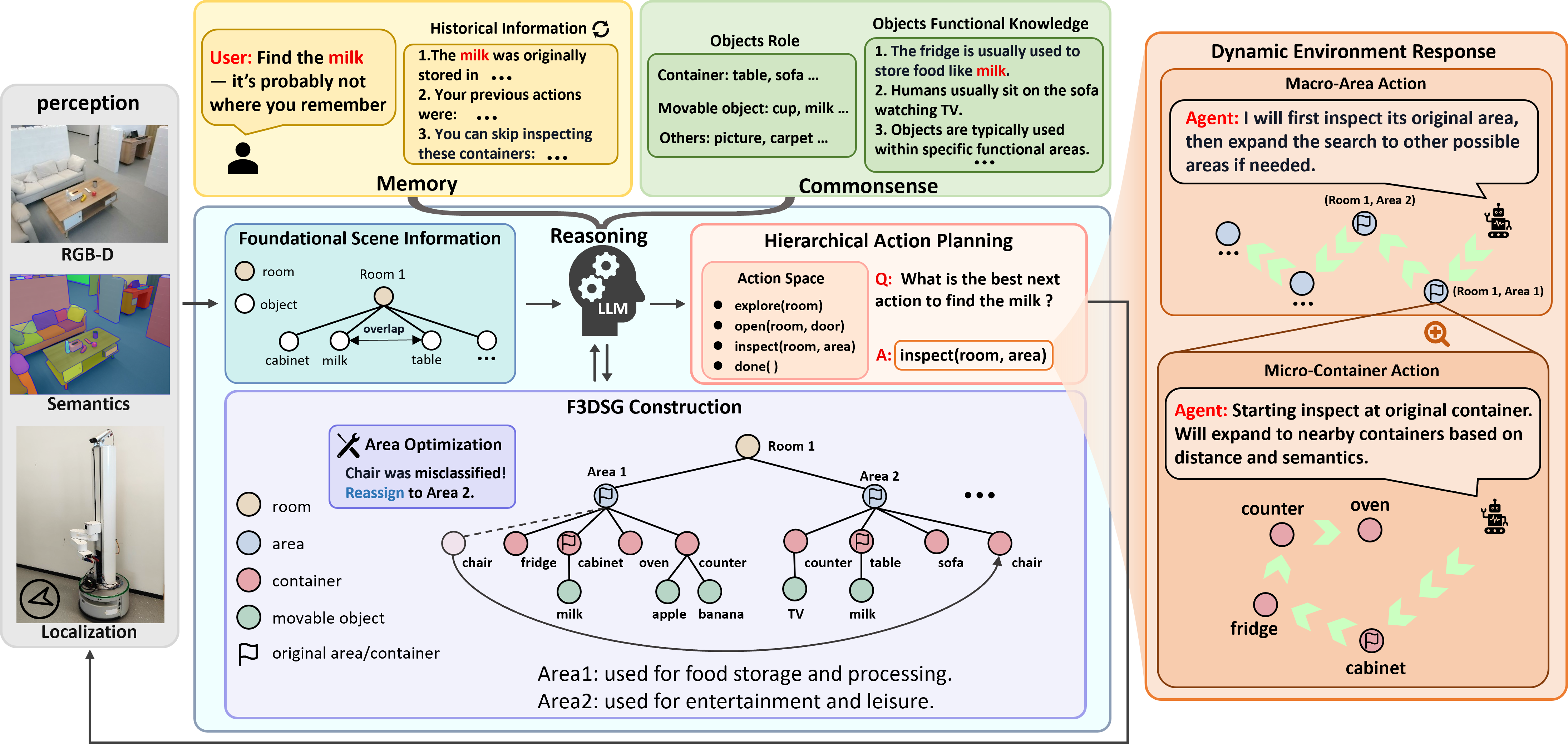}
      \caption{Our CMAR-search framework continuously accumulates environmental memory and leverages commonsense to augment a hierarchical reasoning process across three stages: First, memory and commonsense augment the construction of a F3DSG that organizes the environment into functional areas with their associated containers; Second, F3DSG enables hierarchical action planning at both macro-area and micro-container levels, efficiently identifying and inspecting semantically relevant areas; Third, when environmental dynamics occur (e.g., object relocations), the framework augments relocalization by integrating historical memory with commonsense to prioritize original areas, then inferring alternative search paths based on functional semantics when necessary.}
      \label{fig:pipeline}
   \end{figure*}

Our proposed CMAR-search framework tackles object search in dynamic interactive environments, as shown in Figure~\ref{fig:pipeline}. We assume access to ground truth perception for semantic masks, depth, localization, and handle detection, focusing on the reasoning aspect. The method comprises three components: (1) construction of F3DSG that incorporates functional area segmentation and container-movable object hierarchies; (2) a hierarchical action space for macro-area and micro-container planning; and (3) a dynamic response mechanism leveraging real-time perception, memory, and commonsense.

\subsection{F3DSG Construction}

We first acquire foundational scene information through perception and a door-based room separation approach. This provides the structural basis for hierarchical representations of rooms and objects, along with Voronoi graphs suitable for robot navigation \cite{moma}. Building upon this structural foundation, we further enrich the scene representation through function-aware reasoning to construct the F3DSG using the following method.

\noindent\textbf{Container Layer and Movable Object Layer:}
We classify all objects in the scene (set $\mathcal{O}$) into containers $\mathcal{C}$ and movable objects $\mathcal{M}$ using LLMs. Two objects are considered to have a potential support relationship, and the pair is added to the set $\mathcal{I}$ if they satisfy: (1) overlap on the XY-plane, and (2) are close or overlapping on the Z-axis. If a pair $(c, m)$ satisfies Eq.~\eqref{eq:valid_pair}, it forms a valid container-movable object pair and is included in the set $\mathcal{V}$. All other containers not satisfying this relationship are classified as isolated containers $\mathcal{C}_i$, while remaining movable objects are classified as other objects $\mathcal{S}$.

\begin{equation}
V = \{(c, m) \in I \mid c \in \mathcal{C} \land m \in \mathcal{M}\}
\label{eq:valid_pair}
\end{equation}

   \begin{figure}[t]
      \centering
      \includegraphics[width=\linewidth, keepaspectratio]{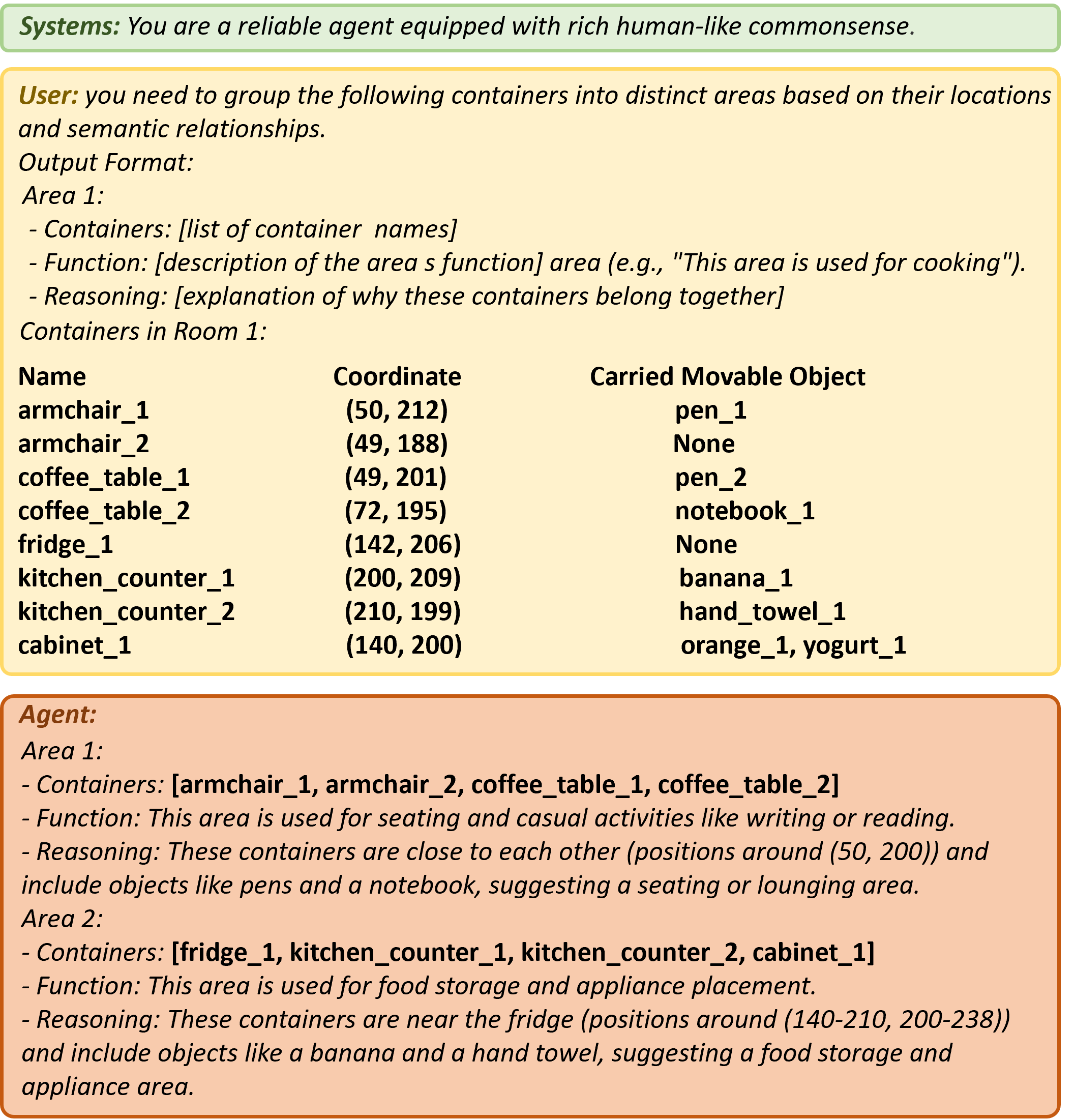}
      \caption{Area Partitioning Prompt: Based on the container names, spatial coordinates, and carried movable objects, the LLMs are instructed to perform area clustering and infer the functional attributes of each area.}
      \label{fig:area_prompt}
   \end{figure}

    \begin{figure}[t]
      \centering
      \includegraphics[width=\linewidth, keepaspectratio]{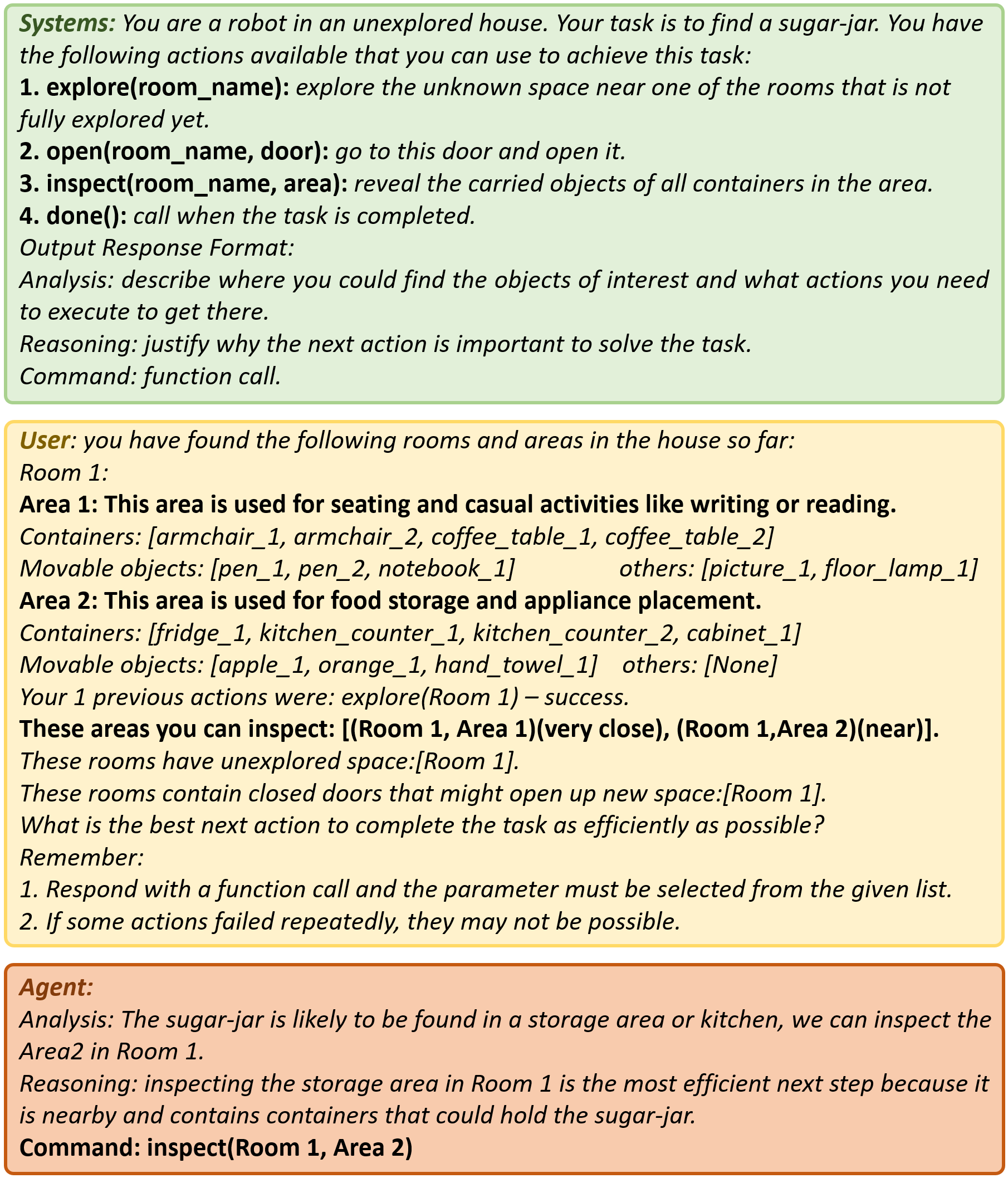}
      \caption{Macro Action Planning Prompt: Provide information about inspectable areas to guide the LLMs in determining whether there are areas that should be inspected or whether exploration-related instructions should be executed.}
      \label{fig:reasoning_prompt}
   \end{figure}

\noindent\textbf{Area Layer:} We input all containers' names, spatial coordinates, and their carried movable objects (using \texttt{None} for isolated containers $\mathcal{C}_i$) to the LLMs. The LLMs then perform area clustering based on both spatial proximity (distance metrics) and semantic relationships, infers each area's functionality, and provides explicit reasoning for this division. The prompt template for this LLMs interaction is shown in Figure~\ref{fig:area_prompt}. Then, we propose an optimization method for LLM-generated area partitions. First, we reduce the number of areas by merging single-container areas to decrease the complexity of subsequent actions planning. Second, we correct positional classification errors by reassigning containers with abnormal spatial distributions to more appropriate areas. The formal implementation follows these rules:
\begin{subequations}\label{eq:region_opt}
\begin{align}
\Phi_1(c) &\coloneqq (|A_i(c)| > 1 \land \exists c' \in A_i(c), d(c,c') < \sigma) \label{eq:phi1} \\
\Phi_2(c) &\coloneqq (|A_i(c)| = 1 \land \forall c', d(c,c') \geq  \sigma) \label{eq:phi2} \\
A_o(c) &= \begin{cases}
A_i(c) & \text{if } \Phi_1(c) \lor \Phi_2(c) \\
\argmin\limits_{A\in\mathcal{A}} \min\limits_{c'\in A} d(c,c') & \text{otherwise} \label{eq:main_result}
\end{cases}
\end{align}
\end{subequations}

\noindent Where $\Phi_1$ and $\Phi_2$ are conditions for preserving initial assignments: 
$\Phi_1$ applies to containers in multi-container areas, and $\Phi_2$ to all areas. 
$A_i(c)$ is the initial area of container $c$, $A_o(c)$ the optimized one, and $|A_i(c)|$ the number of containers in $c$'s area. 
$\mathcal{A}$ is the set of all areas, $d(c,c')$ the Euclidean distance between $c$ and $c'$, and $\sigma$ the distance threshold 
($5\,\mathrm{m}$ in simulation, $3\,\mathrm{m}$ in real-world). 
The $\operatorname{argmin}$ operator selects the area $A \in \mathcal{A}$ containing $c$'s nearest neighbor.
To eliminate the influence of container processing order on optimization results, we adopt a two-phase optimization strategy: the first phase collects optimization decisions for all containers, while the second phase executes all decisions simultaneously. Post-optimization, the function of multi-container areas remains unchanged despite individual container additions or removals. Single-container areas will be deleted if merged into adjacent multi-container areas. When multiple single-container areas merge, the area with the smallest index is retained and its function is set to $\operatorname{None}$. We find that a single iteration is sufficient to achieve optimal outcomes. After optimization, we associate objects in S to their nearest object's corresponding area.

\subsection{Hierarchical Action Planning}

To enable the LLMs to fully leverage CMAR for efficient task planning, we encode the semantic and geometric information from the scene graph into structured text representations \cite{moma}. Our framework features a hierarchical action space that operates at both macro and micro levels, significantly reducing decision complexity and improving search efficiency.

We adopt the method from MoMa-LLM \cite{moma} to construct a bird's-eye-view (BEV) occupancy map and generate a Voronoi graph for navigation based on free space. Frontiers are defined as the boundaries between explored and unexplored regions in the BEV map, guiding the robot's active environmental exploration. For each object in the scene, its closest Voronoi node is also computed to facilitate robot navigation and manipulation. After successfully navigating or opening a container, the robot gains full observability of all objects carried by it. Based on this perceptual capability, we design a macro action space as follows:

\begin{itemize}
\item \textit{explore(\texttt{room\_name})}: Navigates to an unexplored frontier point in the specified room.
\item \textit{open(\texttt{room\_name}, \texttt{door\_name})}: Navigates to a door, opens it, and enters the space behind.
\item \textit{inspect(\texttt{room\_name}, \texttt{area\_name})}: Navigates to each container in the area following the inspection sequence for observation. If the container is closed and operable, performs an opening operation after navigation.
\item \textit{done()}: Terminate the episode when either of the following conditions is met: (1) successful target object localization, or (2) consecutive action execution failures. Throughout operation, the system maintains continuous real-time object detection, immediately terminating the current task upon visual acquisition of the target.
\end{itemize}

We maintain a container blacklist system where, after completing the inspection of a particular area, all containers in that area are added to the blacklist. This mechanism directly supports our hierarchical approach in the inspect action, which decomposes a macro-area command into optimized micro-container operations. For each container within the target area, we compute a composite selection score that balances path distance, semantic relevance, and search risk:

\begin{subequations}\label{eq:container_scoring}
\begin{gather}
\tilde{D}_i = 1 - \frac{D_i - D_{\min}}{D_{\max} - D_{\min}} \label{eq:norm_dist} \\
\tilde{P}_i = \frac{P_i - P_{\min}}{P_{\max} - P_{\min}} \label{eq:norm_prob} \\
S_i = \alpha\tilde{D}_i + \beta\tilde{P}_i + \gamma \prod_{k \neq i}(1-P_k) \label{eq:score} \\
c^* = \mathop{\argmax}\limits_{c_i \in \mathcal{R}} S_i \label{eq:selection}
\end{gather}
\end{subequations}
\noindent where $D_i$ denotes the A* path distance, initially computed from the robot's current position to container $c_i$'s nearest Voronoi node and subsequently calculated between consecutive containers' Voronoi nodes. $P_i$ represents the SBERT-derived \cite{SBERT} cosine similarity between container $c_i$'s name and the target object description. $D_{\max}$ and $D_{\min}$ are the maximum and minimum A* distances among remaining containers $\mathcal{R}$, defined as all containers excluding blacklisted containers and previously selected containers. $P_{\max}$ and $P_{\min}$ are the extreme SBERT similarity values within $\mathcal{R}$; while $\tilde{D}_i$ and $\tilde{P}_i$ are their normalized counterparts in $[0,1]$. The term $\prod_{k \neq i}(1-P_k)$ represents the joint probability that the target is not present in any other remaining container except $c_i$, serving as a risk factor that encourages exploration of containers which, if skipped, might lead to complete search failure. $S_i$ is the composite selection score; $\alpha,\beta,\gamma$ are weighting coefficients ($\alpha=0.4, \beta=0.4, \gamma=0.2$); with $c^*$ being the optimal container selected at each step.

We employ a hierarchical ``room-area'' scene representation to organize information, where each area is described in detail including its functional attributes, containers, movable objects, and other objects. To improve planning efficiency, the system automatically filters out areas where all containers are blacklisted. Since areas are dynamically constructed, our action history only retains two types of operational records: \textit{explore} and \textit{open}. For LLMs decision-making, we constrain the action selection space with prompts like ``These areas you can inspect:'' , while providing proximity information between the robot's current position and each area. This distance is calculated using the A* path distance between the robot and the nearest container in each area. Additionally, the system incorporates other output constraints including chain-of-thought reasoning, with complete prompt examples shown in Figure~\ref{fig:reasoning_prompt}.

\subsection{Dynamic Environment Response}

We propose a human-object-interaction-inspired heuristic target search method to address the challenge of object search in partially explored, dynamic environments. Unlike traditional settings where the environment is assumed to be static and fully known, we focus on more realistic scenarios where objects may be relocated after the initial exploration, and the robot must efficiently re-find them based on limited prior knowledge. Motivated by the observation that movable objects in daily life typically remain within specific functional areas (e.g., a TV remote often moves between the table and sofa in the relaxation area), we develop a hierarchical processing pipeline using CMAR that operates at both area and container levels to enable efficient and robust target relocalization.

\noindent\textbf{Area Level:} The robot leverages its environmental memory to first locate the target object’s initial area within the F3DSG and navigates to that area. For multiple candidate areas, the system prioritizes inspection based on the A* path distance from the robot to the nearest container in each area. If the target is not found in these initial areas, the system employs the LLMs to infer potential relocation areas based on commonsense knowledge of human object usage habits.

\noindent\textbf{Container Level:} When the target area contains original container(s) where the object was previously stored, our system prioritizes inspecting these containers. During inspection, if any changes in movable objects placed on containers are detected, the F3DSG is dynamically updated to reflect the current scene configuration. For the remaining containers beyond the initial ones, the inspection order follows the optimization criteria defined in Eq. (\ref{eq:container_scoring}) ensure efficient search.

\section{EXPERIMENTS}

\subsection{Experiment Setup} 
\textbf{Baselines.} We employ DeepSeek-V3 as our language model for its practical performance-cost balance. We compare our approach against heuristic-based, recent learning-based, and language-based methods.

\noindent \textit{Random:} Uniform random selection among available actions (frontiers and closed objects).

\noindent \textit{Greedy:} Selects the closest actionable target via A* planning in the explored map.

\noindent \textit{ESC-Interactive:} Extends ESC by scoring both frontiers and operable objects using object-room co-occurrences from a fine-tuned language model \cite{27}.

\noindent \textit{HIMOS:} Hierarchical RL approach adapted to large scenes by restricting navigation to target/articulated objects \cite{7}.

\noindent \textit{Unstructured LLM:} Provides raw JSON scene graphs to LLM without structured encoding (adapted from SayPlan \cite{3DSG4}).

\noindent \textit{MoMa-LLM:} Current state-of-the-art using dynamic scene graphs with LLM-grounded room segmentation and open-vocabulary classification \cite{moma}.

\noindent \textit{MoMa-LLM w/ Hydra:} Variant using Hydra's geometric constriction method \cite{3DSG2} instead of door-based segmentation.

\textbf{Metrics.} We use the following metrics to evaluate the performance of our method in scene understanding and multiple dynamic interactive object search tasks.

\noindent \textit{Accuracy:} The proportion of correctly predicted containers among all actual containers, measuring overall prediction correctness.

\noindent \textit{Precision:} The proportion of predicted containers that are true containers, measuring the reliability of predictions.

\noindent \textit{Recall:} The proportion of actual containers that are successfully predicted, evaluating the ability to identify all containers.

\noindent \textit{F1-Score:} The harmonic mean of precision and recall, providing a balanced measure that accounts for both false positives and false negatives.

\noindent \textit{Voronoi Graph Purity:} A measure of the alignment between Voronoi graph components and ground-truth room segmentation, defined as:
\begin{equation}
\text{purity}(\Omega, \mathbb{C}) = \frac{1}{N}\sum_{k}\max_j |\omega_k \cap c_j|
\end{equation}
where $\Omega = {\omega_1, \omega_2, \ldots, \omega_K}$ denotes the Voronoi components, $\mathbb{C} = {c_1, c_2, \ldots, c_J}$ represents the ground-truth rooms, and $N$ is the total number of nodes. The $\max$ operation identifies the best-matching room for each component \cite{moma}.

\noindent \textit{Room Classification Accuracy:} The accuracy of predicting room types based on object distributions within Voronoi components, compared against ground-truth labels \cite{moma}.

\noindent \textit{Object Interactions (OI):} The average number of opening operations performed by the agent per episode.

\noindent \textit{Distance Traveled (DT):} The average distance in meters traveled by the agent per episode from its initial position until the target object is located.

\noindent \textit{Success Rate (SR):} The proportion of episodes in which the agent successfully finds the target object.

\noindent \textit{Success Weighted by Path Length (SPL):} A metric that evaluates efficiency by comparing the traveled path to the shortest possible path, weighted by success. It does not account for the cost of object interactions \cite{SPL}.

\begin{table}[t]
\centering
\caption{Container Classification Evaluation}
\label{tab:container-accuracy}
\begin{tabular}{l | c c c c }
\toprule
Approach & Accuracy ↑ & Precision ↑ & Recall ↑ & F1-Score ↑\\ 
\midrule
OpenIN & 0.648 & 0.972 & 0.648 & 0.778\\
Ours & \textbf{0.957} & \textbf{0.982} & \textbf{0.957} & \textbf{0.969} \\
\bottomrule
\end{tabular}
\end{table}

\begin{table}[t]
\centering
\caption{Room Segmentation Evaluation}
\label{tab:room-segmentation}
\begin{tabular}{l | c c }
\toprule
Approach & \makecell{Voronoi graph \\ purity ↑} & \makecell{Room Classification \\ Accuracy ↑} \\ 
\midrule
Hydra & 0.562 & - \\
MoMa-LLM & 0.615 &  0.276 \\
Ours & \textbf{0.911} & \textbf{0.593} \\
\bottomrule
\end{tabular}
\end{table}

\subsection{Simulation Experiments}
We instantiate the task in the iGibson simulator \cite{igibson} using a Fetch robot, following the iGibson challenge's data split with eight training scenes for module development and prompt engineering, and seven test scenes. In the scene, objects are categorized as movable and non-movable. During scene initialization, movable objects are placed on non-movable objects that meet volume requirements and exhibit certain semantic relevance \cite{moma}. In the dynamic environment response phase, only movable objects are selected as targets, and their new positions are constrained to other semantically and physically compatible non-movable objects.

\noindent\textbf{Scene Understanding:} 
We evaluated the performance of the proposed method in scene understanding across all test scenarios. Using all non-movable objects in the dataset as ground truth for the container category (after excluding special objects such as pictures and windows), we compared the container/non-container classification results obtained by our method and the OpenIN's approach \cite{3DSG1}. As shown in TABLE~\ref{tab:container-accuracy}, our method achieves superior performance across all four evaluation metrics, validating the accuracy of our container-layer construction.

Since there is no standardized evaluation method for our area segmentation (Area Layer construction), we evaluate our approach's scene understanding capability by comparing it with two room segmentation baselines: Hydra's geometry-based method \cite{3DSG2} that detects narrow passages for segmentation, and MoMa-LLM's semantic approach \cite{moma} that uses door positions as partitioning boundaries. Our method adopts a two-stage approach: first, we perform initial room segmentation using door positions similar to MoMa-LLM; then, for rooms that are potentially under-segmented (containing over five containers, typically indicating open-layout spaces), we adapt our Area Layer construction method with modified prompts to further subdivide them into smaller rooms and infer their categories based on spatial proximity and semantic relationships of containers. Each Voronoi node is then connected to its nearest container area node, constructing the separated Voronoi graph for segmentation and classification evaluation. As TABLE~\ref{tab:room-segmentation} demonstrates, our approach surpasses both baselines in Voronoi purity and area classification accuracy, effectively solving the under-segmentation problem that baseline methods encounter in open-layout spaces. 
\begin{table}[t]
\centering
\caption{Interactive Object Search Results}
\label{tab:Interactive Object Search}
\begin{tabular}{l | c c | c c }
\toprule
Model & SR ↑ & SPL ↑ & OI ↓ & DT ↓ \\ 
\midrule
Random & 93.1 & 50.2 & 5.7 & 32.9 \\
Greedy & 85.7 & 50.9 & 8.1 & 22.3 \\
\midrule
ESC-Interactive & 95.4 & 62.7 & 4.1 & 19.6 \\
HIMOS & 93.7 & 48.5 & 4.8 & 35.9 \\
Unstructured LLM & 86.3 & 59.4 & 3.6 & 18.5 \\
MoMa-LLM w/ Hydra & 92.0 & 61.9 & \underline{2.7} & 12.9 \\
MoMa-LLM & 97.7 & 63.6 & 3.9 & 18.2 \\
Ours & \textbf{98.9} & \textbf{79.0} & 2.9 & \underline{9.8} \\
\midrule
MoMa-LLM w/ F3DSG & 94.9 & 76.1 & \textbf{2.1} & \textbf{8.9} \\
Ours w/o area optimization & \textbf{98.9} & 75.1 & 3.2 & 12.8 \\
Ours w/ random order & \textbf{98.9} & \underline{77.8} & 3.0 & 10.6 \\
Ours w/ greedy order & \underline{98.3} & 74.9 & 3.1 & 11.5 \\
\bottomrule
\end{tabular}
\end{table}

\noindent\textbf{Interactive Object Search:} 
The task involves locating target objects in completely unexplored interactive environments. For each test scene, we evaluate the agent through 25 procedurally generated episodes with randomized initial positions, target objects, and object distributions. As shown in TABLE~\ref{tab:Interactive Object Search}, our method demonstrates significant improvements over all baseline approaches, achieving a 24.2\% higher SPL while maintaining competitive SR. Although our hierarchical action planning requires opening all operable containers during area inspection, the precision of our functional area judgment confines this process to a minimal set of candidate regions. This design yields fewer total object interactions than most baselines, demonstrating a superior balance between comprehensive search and operational efficiency. To further validate the superiority of our scene representation, we substituted MoMa-LLM's scene graph with our F3DSG while keeping other components unchanged. This replacement not only maintained the high SR but also achieved significant improvements in SPL, OI, and DT, unequivocally demonstrating that F3DSG enhances exploration efficiency. In ablation studies, we removed the area optimization module and replaced the inspection order with random and greedy strategies (always selecting the nearest container at each step). These variants exhibit degraded performance in SPL, OI, and DT.

\noindent\textbf{Dynamic Environment Response:} 
We design two tasks to evaluate the proposed method's adaptability in dynamic environments. Novel Target Search (NTS): a target object with variable locations is randomly generated in a partially known scene, testing the system's ability to discover and localize new objects under partial prior knowledge. Target Relocalization (TR): a known object is selected from the robot's memory, its position is changed, and the robot must relocate it, assessing its perception of environmental changes and memory updating. To comprehensively evaluate robustness, we define three levels based on the known information of interactive container contents: Level 1 (Fully Unknown), where container contents are completely unknown; Level 2 (Semi-Known), where half of the container contents are known; and Level 3 (Fully Known), where all container contents are known, simulating scenarios with varying degrees of observable content. For experimental control, we assume full exploration of non-interactive parts of the scene, though our method applies to any exploration level.
\begin{table}[t]
\centering
\caption{Novel Target Search Results}
\label{tab:Novel Target Search Results}
\resizebox{\linewidth}{!}{
\begin{tabular}{l | c c | c c | c c | c c}
\toprule
Model & \multicolumn{2}{c|}{Level 1} & \multicolumn{2}{c|}{Level 2} & \multicolumn{2}{c|}{Level 3} & \multicolumn{2}{c}{Average} \\
 & SR & SPL & SR & SPL & SR & SPL & SR & SPL \\
\midrule
MoMa-LLM  &  70.5 & 61.3 & 72.4 & 63.6 & 55.2 & 51.8 & 66.0 & 58.9 \ \\
Ours   & \textbf{97.1} & \underline{77.9} & \textbf{95.2} & \textbf{80.8} & \textbf{91.4} & \textbf{81.2} & \textbf{94.6}& \textbf{80.0}\\
\midrule
MoMa-LLM w/ F3DSG   &85.7 & 75.5 & 74.3 & 65.6 & 61.9 & 53.2 & 74.0& 64.8\\
Ours w/o area optimization & 94.3 & \textbf{78.9} & \underline{91.4} & 79.6 & \underline{88.6} & \underline{76.4} &91.4 &\underline{78.3} \\
Ours w/ random order &  \underline{96.2} & 76.1 & \textbf{95.2} & 78.7 & \textbf{91.4} & 72.8 &\underline{94.3} &75.9\\
Ours w/ greedy order & \textbf{97.1} & 74.3 & \textbf{95.2} & \underline{79.8} & \textbf{91.4} & 71.6 & \textbf{94.6}&75.2 \\
\bottomrule
\end{tabular}%
}
\end{table}

\begin{table}[t]
\centering
\caption{Target Relocalization Results}
\label{tab:Target Relocalization Results}
\resizebox{\linewidth}{!}{
\begin{tabular}{l | c c | c c | c c | c c}
\toprule
Model & \multicolumn{2}{c|}{Level 1} & \multicolumn{2}{c|}{Level 2} & \multicolumn{2}{c|}{Level 3} & \multicolumn{2}{c}{Average} \\
 & SR & SPL & SR & SPL & SR & SPL & SR & SPL \\
\midrule
MoMa-LLM  & 81.0 & 71.2 & 85.7 & 76.8 & 84.8 & 72.2 & 83.8 & 73.4 \\
Ours      & \textbf{98.1} & \textbf{84.9} & \textbf{98.1} & \textbf{86.4} & \underline{97.1} & \textbf{88.3} & \textbf{98.1} & \textbf{86.5} \\
\midrule
MoMa-LLM w/ F3DSG           & 88.6 & 73.0 & 88.6 & 75.5 & 86.7 & 72.0 & 88.0 & 73.5 \\
Ours w/o area optimization       & 96.2 & \underline{82.4} & \underline{97.1} & 77.2 & \textbf{98.1} & 87.8 & 97.1 & \underline{82.5} \\
Ours w/ random order            & \underline{97.1} & 79.2 & \underline{97.1} & 78.9 & \textbf{98.1} & \underline{88.1} & 97.4 & 82.1 \\
Ours w/ greedy order            & \textbf{98.1} & 78.1 & \textbf{98.1} & \underline{81.2} & 96.2 & 82.8 & \underline{97.5} & 80.7 \\
\bottomrule
\end{tabular}%
}
\end{table}

For each scene and level, we conduct 15 episodes with randomized initial knowledge. We use MoMa-LLM \cite{moma} as the baseline, where for the TR task, the robot first navigates to the original target location and upon detecting a change, applies MoMa-LLM to reason about the subsequent action. Results for NTS and TR are shown in TABLE~\ref{tab:Novel Target Search Results} and TABLE~\ref{tab:Target Relocalization Results}, respectively. Our method demonstrates substantial improvements over the baseline, achieving 43.3\% higher SR and 35.8\% higher SPL in NTS tasks, and 17.1\% higher SR and 17.8\% higher SPL in TR tasks. Failure analysis reveals that MoMa-LLM lacks systematic environment organization, relying on navigating to objects potentially related to the target, leading to excessive attempts and significantly reduced SR and SPL. In some cases, it prematurely terminates search upon concluding the target is unreachable. Replacing its scene representation with F3DSG improves performance, but limitations in the action space still prevent thorough inspection of specific areas. Our method leverages the CMAR mechanism to effectively improve the efficiency of both NTS and TR by fully utilizing memory and object usage commonsense. Ablation studies further validate the contribution of each component to the overall performance.

\begin{table}[t]
\centering
\caption{Real-World Experiments Results}
\label{tab:Real-World Experiments Results}
\begin{tabular}{l | c c | c c | c c }
\toprule
Model & \multicolumn{2}{c|}{IOS} & \multicolumn{2}{c|}{NTS} & \multicolumn{2}{c}{TR}  \\
 & OI & DT & OI & DT & OI & DT  \\
\midrule
MoMa-LLM  & 1.8 & 12.6 & 1.5 & 10.5 & 1.9 & 12.1  \\
Ours      & \textbf{1.1} & \textbf{8.3} & \textbf{0.9} & \textbf{5.3} & \textbf{0.8} & \textbf{6.7} \\
\bottomrule
\end{tabular}%
\end{table}

\begin{figure}[t]
  \centering
  \includegraphics[width=\linewidth, keepaspectratio]{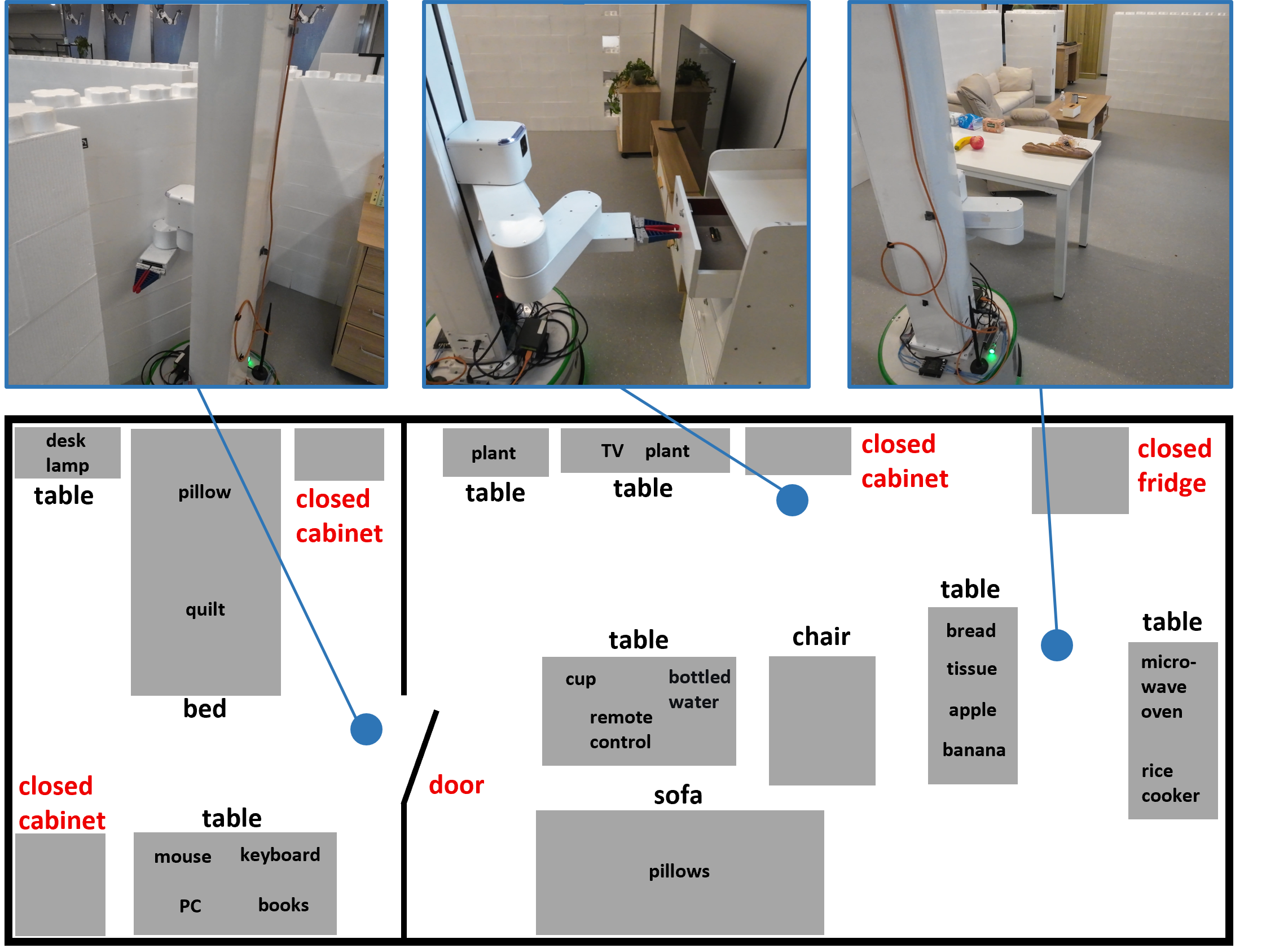}
  \caption{Real-world household experimental environment. Interactive objects are labeled in red.}
  \label{fig:realworld}
\end{figure}

\subsection{Real-World Experiments}
We construct a real-world household environment with two distinct rooms, as illustrated in Figure~\ref{fig:realworld}: an open-layout room combining the kitchen and living room functions (with no internal doors), and a separate bedroom. The visible regions of the scene were first mapped using an existing method \cite{3DSG5} to extract object information. In the Interactive Object Search (IOS) task, the robot was incrementally provided with map information from areas covered by its camera during operation. For both NTS and TR tasks, the system is granted complete information about non-interactive parts of the scene (equivalent to Level 1 in simulation, where container contents are completely unknown).

We perform 10 episodes for each task category, including open-vocabulary instructions such as ``I am hungry, find me some food'' and ``I am thirsty, find me a drink''. An episode is considered successful if any object relevant to the instruction enters the robot's field of view. MoMa-LLM \cite{moma} is employed as the baseline method. Owing to the constrained scale of the environment and limited number of objects, both our method and the baseline achieve perfect success rates across all tasks. We therefore evaluate exploration efficiency using two metrics: the number of Object Interactions and the Distance Traveled. As summarized in the accompanying TABLE~\ref{tab:Real-World Experiments Results}, our method consistently outperforms the baseline in all three types of tasks. The advantage is particularly evident in NTS and IOS tasks, where our CMAR-search framework effectively utilizes prior knowledge to efficiently adapt to real-world dynamic variations.

\section{CONCLUSIONS}

We propose CMAR-search, a novel framework for dynamic interactive object search that augments robotic reasoning through the synergistic integration of memory and commonsense. Our method constructs an F3DSG using LLM-based functional area inference and a hierarchical action space to enable efficient macro-area and micro-container planning. The CMAR mechanism integrates real-time perception, historical memory, and object-functional commonsense to dynamically adapt to environmental changes. Extensive experiments in simulation and real-world settings demonstrate that our approach outperforms baselines in scene understanding, success rate, efficiency, and robustness, providing a scalable solution for long-horizon interactive search tasks in dynamic human environments.





\section*{ACKNOWLEDGMENT}

This work was supported by the Shenzhen Science and Technology Program [Grant No. CJGJZD20240729141702003] and the Beijing Institute of Computer Technology and Applications [Grant No. 2023WDZC02002]. The authors would also like to thank Shenzhen Pudu Technology Co., Ltd. for the support and resources provided, which were essential to the completion of this research. Finally, the authors acknowledge the use of GPT-4 for language polishing and grammatical corrections during the preparation of this manuscript.


\bibliographystyle{ieeetr}
\bibliography{references}  

\end{document}